%% file: main.tex
\documentclass[lettersize,journal]{IEEEtran}
\usepackage{amsmath,amsfonts}
\usepackage{algorithmic}
\usepackage{algorithm}
\usepackage{array}
\usepackage[caption=false,font=normalsize,labelfont=sf,textfont=sf]{subfig}
\usepackage{textcomp}
\usepackage{stfloats}
\usepackage{url}
\usepackage{verbatim}
\usepackage{graphicx}
\usepackage{cite}
\usepackage{svg}
\usepackage{bbding} 
\usepackage{mathrsfs}
\usepackage{booktabs}

\usepackage{makecell}
\usepackage{float}
\usepackage{color}
\usepackage{dsfont}
\usepackage{bbm}

\begin{document}


\title{StabStitch++: Unsupervised Online Video Stitching with Spatiotemporal Bidirectional Warps}

\author{Lang Nie, Chunyu Lin, Kang Liao, Yun Zhang, Shuaicheng Liu,~\IEEEmembership{Senior Member,~IEEE}, \\Yao Zhao,~\IEEEmembership{Fellow,~IEEE}
		
\thanks{Lang Nie, Chunyu Lin, and Yao Zhao are with the Institute of Information Science, Beijing Jiaotong University, Beijing 100044, China, and also with Visual Intelligence +X International Cooperation Joint Laboratory of MOE, Beijing 100044, China (e-mail: nielang@bjtu.edu.cn, cylin@bjtu.edu.cn, yzhao@bjtu.edu.cn).}
\thanks{Kang Liao is with the School of Computing and Data Science, Nanyang Technological University, Singapore (e-mail:  kang.liao@ntu.edu.sg).}
\thanks{Yun Zhang is with the School of Media Engineering, Communication University of Zhejiang, Hangzhou 310018, China (e-mail: zhangyun@cuz.edu.cn).}
\thanks{Shuaicheng Liu is with the School of Information and Communication Engineering, University of Electronic Science and Technology of China, Chengdu 611731, China (e-mail: liushuaicheng@uestc.edu.cn).}
 \thanks{This work was supported by the National Natural Science Foundation of China (NSFC) under Grants U2441242 and 62172032, as well as by the Open Fund of Zhejiang Key Laboratory of Film and TV Media Technology. (Corresponding author: Chunyu Lin.)\\
 }
}

\markboth{Journal of \LaTeX\ Class Files,~Vol.~14, No.~8, August~2021}%
{Shell \MakeLowercase{\textit{et al.}}: A Sample Article Using IEEEtran.cls for IEEE Journals}


\maketitle

\input{sections/Abstract.tex}
		
 


\input{sections/Introduction.tex}

	\input{sections/Relatedwork.tex}
	\input{sections/Methodology.tex}
     \input{sections/Online.tex}

\input{sections/Experiment.tex}
    \input{sections/Limitation.tex}
	\input{sections/Conclusion.tex}

    \bibliographystyle{IEEEtran}
	\bibliography{bib}

\vfill

\end{document}

%% file: sections/Abstract.tex
\begin{abstract}
\label{sec:Abstrat}
We retarget video stitching to an emerging issue, named \textit{\textbf{warping shake}}, which unveils the temporal content shakes induced by sequentially unsmooth warps when extending image stitching to video stitching. Even if the input videos are stable, the stitched video can inevitably cause undesired warping shakes and affect the visual experience. 
To address this issue, we propose \textit{\textbf{StabStitch++}}, a novel video stitching framework to realize spatial stitching and temporal stabilization with unsupervised learning simultaneously. 
First, different from existing learning-based image stitching solutions that typically warp one image to align with another, we suppose a virtual midplane between original image planes and project them onto it. Concretely, we design a differentiable bidirectional decomposition module to disentangle the homography transformation and incorporate it into our spatial warp, evenly spreading alignment burdens and projective distortions across two views.
Then, inspired by camera paths in video stabilization, we derive the mathematical expression of stitching trajectories in video stitching by elaborately integrating spatial and temporal warps. 
Finally, a warp smoothing model is presented to produce stable stitched videos with a hybrid loss to simultaneously encourage content alignment, trajectory smoothness, and online collaboration. Compared with \textit{StabStitch} that sacrifices alignment for stabilization, \textit{StabStitch++} makes no compromise and optimizes both of them simultaneously, especially in the online mode. 
To establish an evaluation benchmark and train the learning framework, we build a video stitching dataset with a rich diversity in camera motions and scenes. 
Experiments exhibit that \textit{StabStitch++} surpasses current solutions in stitching performance, robustness, and efficiency, offering compelling advancements in this field by building a real-time online video stitching system. 
The code and dataset are available at \url{https://github.com/nie-lang/StabStitch2}.
\end{abstract}

%% file: sections/Introduction.tex

\section{Introduction}
\label{sec:Introduction}

\begin{figure*}[!t]
  \centering
  \includegraphics[width=.97\textwidth]{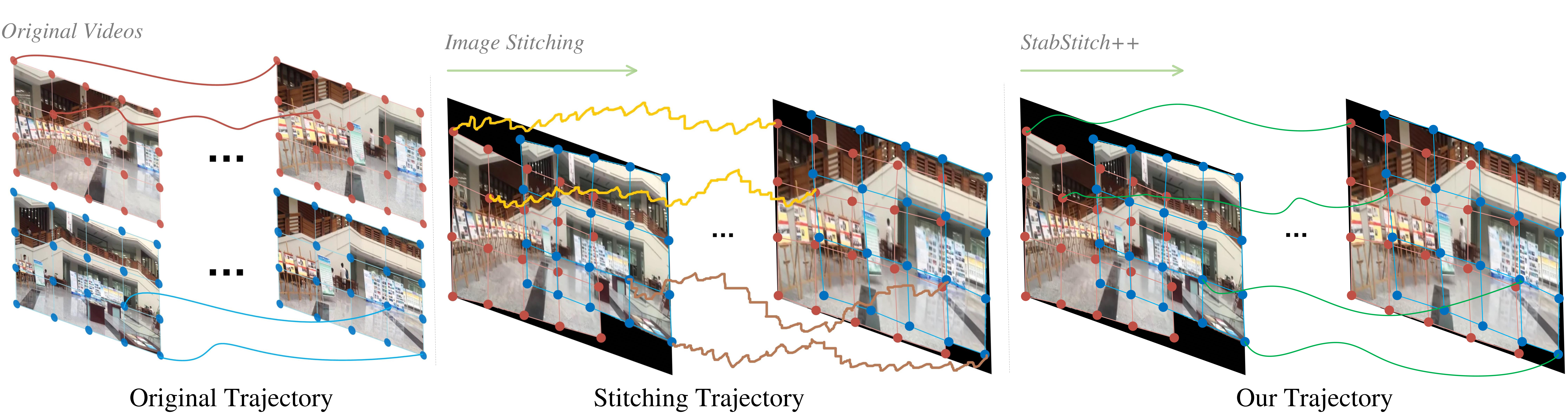}
  \caption{The occurrence and elimination of warping shakes. Left: stable camera trajectories for input videos. Middle: warping shakes are produced by image stitching, yielding unsmooth stitching trajectories. Right: the proposed \textit{StabStitch++} eliminates these shakes successfully.}
  \label{fig:fig1}
\end{figure*}

\IEEEPARstart{V}{ideo} stitching techniques are commonly employed to create panoramic or wide field-of-view (FoV) displays from different viewpoints with limited FoV. Due to their practicality, they are widely applied in autonomous driving \cite{lai2019video}, video surveillance \cite{liu2018future}, virtual reality \cite{lo2021efficient}, etc.
Our work lies in the most common and challenging case of video stitching with hand-held cameras. It does not require camera poses, motion trajectories, or temporal synchronization. It merges multiple videos, whether from multiple cameras or a single camera, capturing multiple videos to create a more immersive representation of the captured scene. Moreover, it transforms video production into an enjoyable and collaborative endeavor among a group of individuals.

Compared with video stitching, image stitching has been studied more extensively and profoundly, which inevitably throws the question of whether existing image stitching solutions can be directly extended to video stitching. Pursuing this thought, we initially leverage existing image stitching algorithms \cite{jia2021leveraging}\cite{nie2023parallax} to process hand-held camera videos. Subsequently, we observe that although the stitched results for individual frames are remarkably natural, there is apparent content jitter among temporally consecutive frames. It is also important to note that the jitter effect does not originate from the inherent characteristics of the source video itself, although these videos are captured by hand-held cameras. In fact, due to the advancements and widespread adoption of video stabilization in both hardware and software nowadays, the source videos obtained from hand-held cameras are typically stable unless deliberately subjected to shaking. For clarity, we define such content jitter as \textit{warping shake}, which describes the temporal content instability induced by temporally unsmooth warps, irrespective of the stability of source videos. Fig. \ref{fig:fig1} (left and middle) illustrates the occurrence process of warping shakes.

Existing video stitching solutions \cite{nie2017dynamic}\cite{su2016video}\cite{guo2016joint}\cite{lin2016seamless}\cite{jiang2015video} follow a strong assumption that each source video from freely moving hand-held cameras suffers from heavy and independent shakes. Consequently, every source video necessitates stabilization via warping, contradicting the current prevalent reality that video stabilization technology has already been widely integrated into various portable devices (\textit{e.g.}, cellphones, DV cameras, and UAVs). In addition, these approaches, to jointly optimize video stabilization and stitching, often establish a sophisticated non-linear solving system consisting of various energy terms. To find the optimal parameters, an iterative solving strategy is typically employed. Each iteration involves several steps dedicated to optimizing different parameters separately, resulting in a rather slow inference speed. The complicated optimization procedures also impose stringent requirements on the quality of input videos (\textit{e.g.}, sufficient, accurate, and evenly distributed matching points), making video stitching systems fragile and not robust in practical applications.

To solve the above issues, we present a novel unsupervised online video stitching framework (termed \textit{StabStitch++}) to simultaneously realize spatial stitching and temporal stabilization. 
First, existing learning-based image stitching solutions typically warp one image to align with another, concentrating all the alignment challenges and projective distortions into a single view. In contrast, our spatial warp introduces a differentiable bidirectional decomposition module to evenly distribute the burdens across both views, thereby boosting alignment and reducing distortions.
It determines a virtual midplane by disentangling the global homography transformation and then carries out local spatial stitching by projecting the original image planes onto it.
Then, building upon the current condition that source videos are typically stable, we simplify this task to stabilize the warped videos by removing warping shakes as illustrated in Fig. \ref{fig:fig1} (middle and right). To get the shaked trajectories, we derive the mathematical formulation of stitching trajectories in the warped video from the experience of camera paths in video stabilization by ingeniously combining spatial and temporal warps. 
Finally, to get stable stitched videos, we propose a warp smoothing model to simultaneously encourage content alignment, trajectory smoothness, and online collaboration within a hybrid loss. It is worth mentioning that this joint optimization aims to find the optimal solution that satisfies all of these conditions simultaneously, rather than sacrificing one condition to enhance the others.

Diverging from conventional offline video stitching approaches that require complete videos as input, \textit{StabStitch++} stitches and stabilizes the current online frame to composite a high-quality stitched video only with historical frames. Besides, its efficient designs further contribute to a real-time online video stitching system with minimum latency.

As there is no proper dataset readily available, we build a holistic video stitching dataset to train the proposed framework. Moreover, it could serve as a comprehensive benchmark with rich camera motions and scene diversity to evaluate image/video stitching methods. In summary, the main contributions of this paper are concluded as follows:
\begin{itemize}
	\item We retarget video stitching to an emerging issue, termed \textit{warping shake}, and reveal its occurrence when extending image stitching to video stitching.  
	\item We present \textit{StabStitch}, the first online video stitching framework, with a pioneering step to integrate video stitching and stabilization with unsupervised learning.   
    \item We propose a holistic video stitching benchmark dataset with diverse scenes and camera motions, which we hope can promote other related research work.  
    \item Compared with state-of-the-art image/video stitching solutions, our method achieves superior performance in terms of scene robustness, inference speed, and stitching/stabilization effect.  
\end{itemize}

In comparison to our previous conference version \cite{nie2024eliminating}, we
make the following new contributions substantially:
\begin{itemize}
	\item We propose a differentiable bidirectional decomposition module to carry out bidirectional warping on a virtual middle plane, evenly spreading warping burdens across both views. 
    It benefits both image and video stitching, demonstrating universality and scalability.
    \item A new warp smoothing model is presented to simultaneously encourage content alignment, trajectory smoothness, and online collaboration using a hybrid loss.  Different from our conference version that sacrifices alignment for stabilization, the new model searches for a joint optimum in online mode. 
    \item With the above new contributions, we extend \textit{StabStitch} to \textit{StabStitch++} with better alignment, fewer distortions, and higher stability. It can deal with not only common stable videos but also unstable videos.
\end{itemize}


\vspace{-0.1cm}

%% file: sections/Relatedwork.tex
\section{Related Work}
\label{sec:relatedwork}
Here, we briefly review image stitching, video stabilization, and video stitching techniques, respectively.

\vspace{-0.1cm}
\subsection{Image Stitching}
Traditional image stitching methods usually detect keypoints \cite{lowe2004distinctive} or line segments \cite{von2008lsd} and then minimize the projective errors to estimate a parameterized warp by aligning these geometric features. To eliminate the parallax misalignment \cite{zhang2014parallax}, the warp model is extended from global homography transformation \cite{brown2007automatic} to other elastic representations, such as mesh \cite{zaragoza2013projective}, TPS \cite{li2017parallax}, superpixel \cite{lee2020warping}, and triangular facet\cite{li2019local}. Meanwhile, to keep the natural structure of non-overlapping regions, a series of shape-preserving constraints is formulated with the alignment objective. For instance, SPHP \cite{chang2014shape} and ANAP \cite{lin2015adaptive} linearized the homography and slowly changed it to the global similarity to reduce projective distortions; DFW \cite{li2015dual}, SPW \cite{liao2019single}, and LPC \cite{jia2021leveraging} leveraged line-related consistency to preserve geometric structures; GSP \cite{chen2016natural} and GES-GSP \cite{du2022geometric} added a global similarity before stitching multiple images together so that the warp of each image resembles a similar transformation as a whole; etc. Besides, Zhang \textit{et\ al.} \cite{zhang2020content} re-formulated image stitching task with regular boundaries by simultaneously optimizing alignment and rectangling \cite{he2013rectangling}\cite{nie2022deep}.

Recently, learning-based image stitching solutions emerged. They feed the entire images into the neural network, encouraging the network to directly predict the corresponding parameterized warp model (\textit{e.g.}, homography \cite{nie2020view}\cite{nie2022learning}\cite{jiang2022towards}, multi-homography \cite{song2021end}, TPS \cite{nie2023parallax}\cite{kim2024learning}\cite{zhang2024recstitchnet}, and optical flow \cite{kweon2023pixel}\cite{jia2023learning}). Compared with traditional methods based on sparse geometric features, these learning-based solutions train the network parameters to adaptively capture semantic features by establishing dense pixel-wise optimization objectives. They show better robustness in various cases, especially in challenging cases where hand-craft features are few to detect.

\vspace{-0.2cm}
\subsection{Video Stabilization}
Traditional video stabilization can be categorized into 3D \cite{Liu2009content}\cite{liu2012video}, 2.5D \cite{Liu2011subspace}\cite{Goldstein2012video}, and 2D \cite{Matsushita2006full}\cite{Grundmann2011auto}\cite{ma2019effective} methods, according to different motion models. The 3D solutions model the camera motions in 3D space or require extra scene structure for stabilization. The structure is either calculated by structure-from-motion (SfM) \cite{Liu2009content} or acquired from additional hardware, such as a depth camera \cite{liu2012video}, a gyroscope sensor \cite{karpenko2011digital}, or a lightfield camera \cite{smith2009light}. Given the intensive computational demands of these 3D solutions, 2.5D approaches relax the full 3D requirement to partial 3D information. To this end, additional 3D constraints are established, such as subspace projection \cite{Liu2011subspace} and epipolar geometry \cite{Goldstein2012video}. Compared with them, the 2D methods are more efficient with a series of 2D linear transformations (\textit{e.g.}, affine, homography) as camera motions. To deal with large-parallax scenes, spatially varying motion representations are proposed, such as homography mixture \cite{Grundmann2012calibration}, mesh \cite{liu2013bundled}, vertex profile \cite{liu2016meshflow}, optical flow \cite{liu2014steadyflow}, etc. Moreover, some special approaches focus on special scenes and specific input (\textit{e.g.}, selfie \cite{yu2018selfie}\cite{yu2021real}, 360 \cite{kopf2016360}\cite{tang2019joint}, and hyperlapse \cite{joshi2015real} videos).

In contrast, learning-based video stabilization methods directly regress unstable-to-stable transformation from data. Most of them are trained with stable and unstable video pairs acquired by special hardware in a supervised manner \cite{wang2018deep}\cite{xu2018deep}\cite{zhang2023minimum}. To relieve data dependence, DIFRINT \cite{choi2020deep} proposed the first unsupervised solution via neighboring frame interpolation. To get a stable interpolated frame, only stable videos are used to train the network. Different from it, DUT \cite{xu2022dut} established unsupervised constraints for motion estimation and trajectory smoothing, learning video stabilization by simply watching unstable videos. 

\vspace{-0.2cm}
\subsection{Video Stitching}
Video stitching has received much less attention than image stitching. Early works \cite{lai2019video}\cite{perazzi2015panoramic} stitched multiple videos frame-by-frame and focused on the temporal consistency of stitched frames. But the input videos were captured by cameras fixed on rigs. For hand-held cameras with free and independent motions, there is a significant increase in temporal shakes. To deal with it, videos were first stitched and then stabilized in \cite{hamza2015stabilization}, while \cite{lin2016seamless} did it in an opposite way (\textit{e.g.}, videos were firstly stabilized, and then stitched). Both of them accomplished stitching or stabilization in a separate step. Later, a joint optimization strategy was commonly adopted in \cite{su2016video}\cite{guo2016joint}\cite{nie2017dynamic}, where \cite{nie2017dynamic} further considered the dynamic foreground by background identification. However, solving such a joint optimization problem regarding stitching and stabilization is fragile and computationally expensive. To this end, we rethink the video stitching problem from the perspective of warping shake and propose the first (to our knowledge) unsupervised online solution for hand-held cameras.


%% file: sections/Methodology.tex




\begin{figure*}[!t]
  \centering
  \includegraphics[width=.97\textwidth]{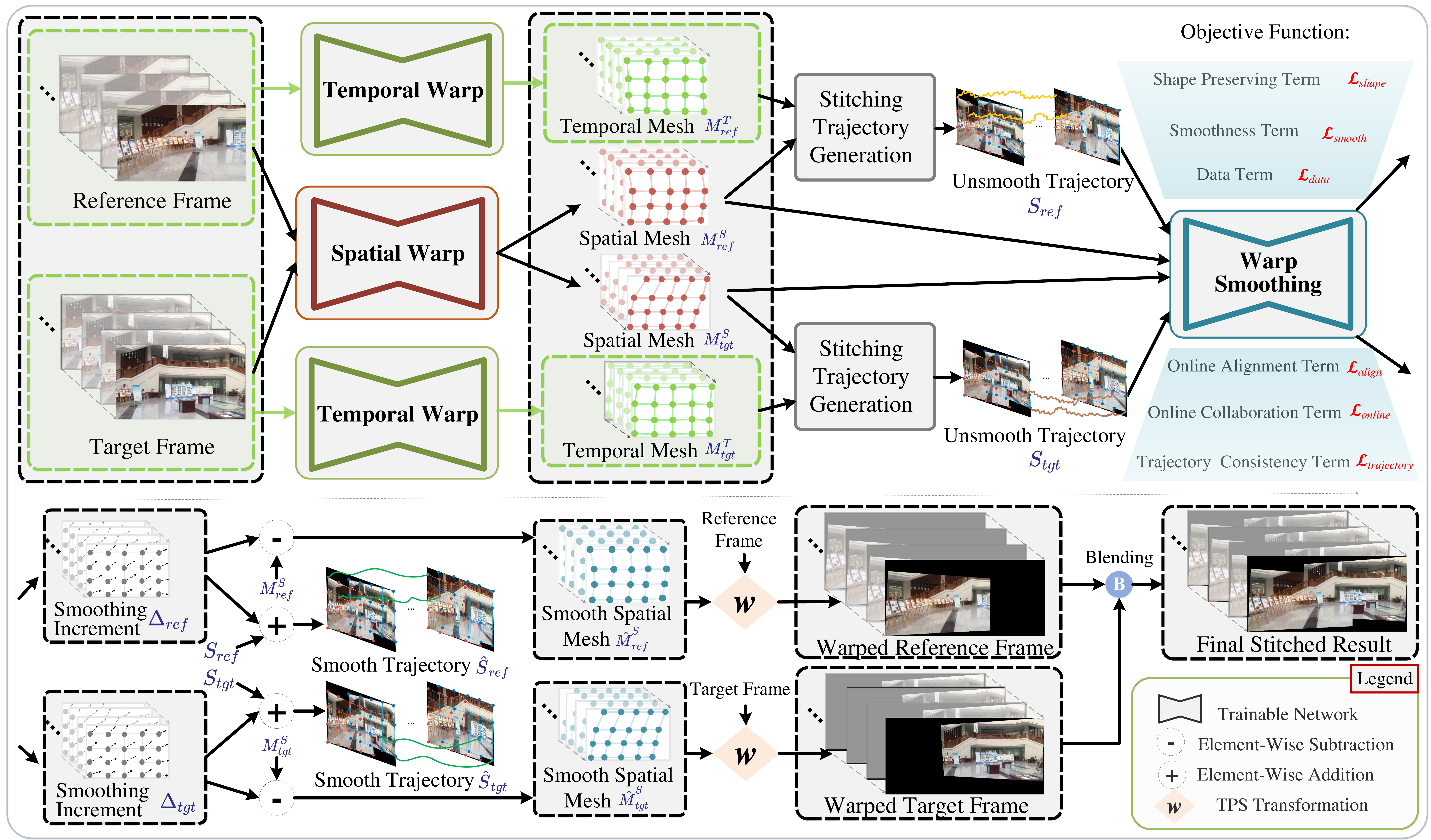}
  \caption{The overview of \textit{StabStitch++}. We first get the spatial and temporal meshes through spatial and temporal warp models. Then stitching trajectories can be derived by integrating spatial and temporal warps. Finally, a warp smoothing model is leveraged to produce stable stitched frames.}
  \label{fig:framework}
  \vspace{-0.4cm}
\end{figure*}

\vspace{-0.1cm}
\section{Methodology}
\label{sec:Methodology}

The framework of \textit{StabStitch++} is illustrated in Fig. \ref{fig:framework}, where we take consecutive video frames (\textit{i.e.}, the reference frames and target frames) as input and output the stable stitched frames. It consists of three trainable warp models: spatial warp, temporal warp, and warp smoothing models. We first introduce our spatial and temporal warp models, of which the spatial warp model includes a bidirectional decomposition module. Then, we derive the expression of stitching trajectories for video stitching by integrating spatial and temporal warps, which implicitly represent the warping shakes.
Thereafter, a warp smoothing model is proposed to produce stable stitched video with smooth stitching paths.

Here, we briefly review the preliminaries of UDIS++~\cite{nie2023parallax} to distinguish the unidirectional warps from ours. UDIS++ employs a two-stage warping process: the first stage estimates a global homography transformation~\cite{hartley2003multiple}, while the second stage refines it using local Thin-Plate Spline (TPS) transformation~\cite{bookstein1989principal}. TPS is a principal warp \cite{nie2024semi} that describes the deformations specified by finitely many point-correspondences in an irregular spacing between a flat image and a warped one. UDIS++ reparameterizes the two different transformations in a uniform representation for joint learning. Concretely, the homography is parameterized as the motions of four vertices, while the TPS is represented as the motions of a uniform grid of $(U+1)\times (V+1)$ control points predefined across the image. A key limitation of UDIS++ is its reliance on unidirectional warping, where only the target image is warped toward the reference, concentrating distortions in a single view. In our work, we address this limitation through bidirectional decomposition, as detailed in Sec. \ref{spatial_warp}2.

\vspace{-0.2cm}
\subsection{Spatial Warp}
\label{spatial_warp}
\subsubsection{Network}
Given a reference and target image ($I^t_{ref}$ and $I^t_{tgt}$), the spatial warp model aims to estimate the spatial deformation that can naturally align the two images. The network is shown in Fig. \ref{fig:network} (a). Similar to UDIS++~\cite{nie2023parallax}, it has a global-to-local structure to combine homography and TPS in a common network. 
Particularly, the images are first input into a backbone network \cite{he2016deep} with shared weights to capture high-level semantic features.
Then, the global part calculates the global correlation \cite{9605632} with the feature maps of the last layer and regresses the global homography.
Next, we design a bidirectional decomposition module to disentangle the estimated homography $H(t)$ into $H_{ref}(t)$ and $H_{tgt}(t)$. It supposes there is a virtual middle and projects the feature maps of the last but one layer onto it. Finally, we adopt the local correlation layer (\textit{i.e.}, cost volume \cite{sun2018pwc}) to capture short-range matching information and regress residual control point motions. The disentangled homography can be transformed into the representations of control point motions and combined with the local residual motions to get final spatial warps (denoted as the sum of control point motions $m_{ref}^S(t)$, $m_{tgt}^S(t)$).

\subsubsection{Bidirectional Decomposition} 
We intend to find a virtual middle view between the reference and target views. However, it is hard to obtain the camera extrinsics (\textit{i.e.,} translation and rotation) \cite{liao2023deep} due to limited overlapping regions and the lack of intrinsic. To this end, we simplify the image view as a plane and leverage homography to represent the transformation from one plane to another. This planar transformation can be characterized by a $3\times 3$ matrix with eight degrees of freedom: two each for translation, rotation, (an)isotropic scale, and perspectivity. 
To obtain the middle plane, an intuitive approach is to ensure the magnitude of each attribute (\textit{e.g.}, translation, scale, etc.) is halved from the original transformation. But these components are intricately coupled within a $3\times 3$ matrix (\textit{e.g.}, scaling and rotation are intertwined) and the change of their magnitude is non-linear. To this end, we propose to use the 4-pt representation of deep homography \cite{detone2016deep} as an alternative, as outlined below:
\begin{equation}
H(t) = \begin{pmatrix}
 h_{11} & h_{12} & h_{13}\\
 h_{21} & h_{22} & h_{23}\\
 h_{31} & h_{32} &h_{33}
\end{pmatrix} \sim \begin{pmatrix}
 \Delta u_1& \Delta v_1 \\
 \Delta u_2& \Delta v_2 \\
 \Delta u_3& \Delta v_3\\
\Delta u_4& \Delta v_4
\end{pmatrix},
\end{equation}
where ($\Delta u$, $\Delta v$) denotes the motions of four vertices on the image. Using the four vertices and their motions, we can get four pairs of matched points, uniquely determining a homography transformation.

\begin{figure*}[!t]
  \centering
  \includegraphics[width=.97\textwidth]{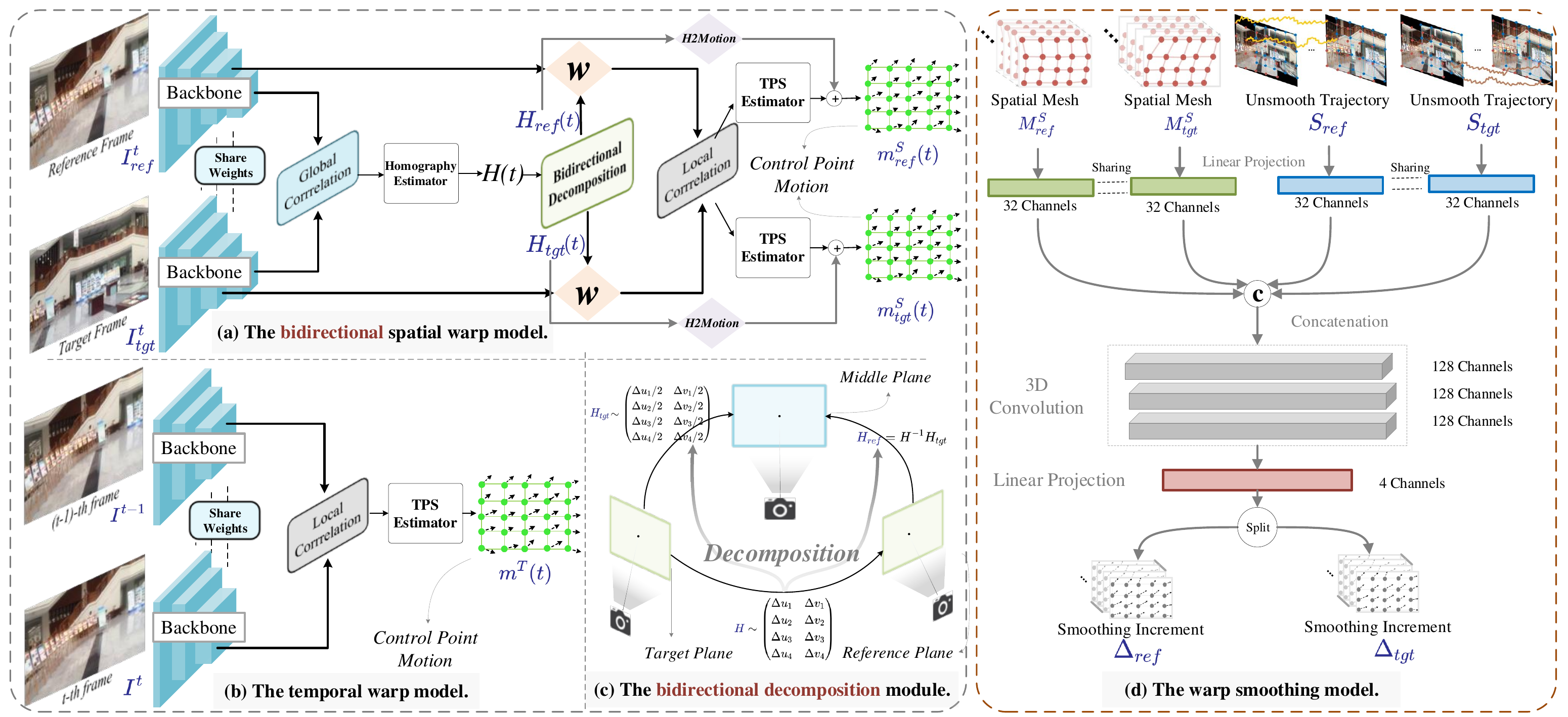}
  \caption{The network structures of our warping models. The spatial warp, temporal warp, and warp smoothing models are depicted in (a)(b)(d).}
  \label{fig:network}
  \vspace{-0.4cm}
\end{figure*}

In the 4-pt representation, all elements can be uniformly interpreted as spatial displacements. Different from the matrix representation, each element is independent and its magnitude change is linear. Based on the above properties, we present to determine the virtual middle plane by halving all the displacements from the original 4-pt representation:
\begin{equation}
H_{tgt}(t) \sim \begin{pmatrix}
 \Delta u_1/2& \Delta v_1/2 \\
 \Delta u_2/2& \Delta v_2/2 \\
 \Delta u_3/2& \Delta v_3/2\\
\Delta u_4/2& \Delta v_4/2
\end{pmatrix}.
\end{equation}
Then, the homography mapped from the reference plane to the middle one is formulated as:
\begin{equation}
H_{ref}(t) = H^{-1}(t)H_{tgt}(t).
\end{equation}
Fig. \ref{fig:network} (c) shows the whole decomposition process and all operations are differentiable, making it easy to incorporate into deep learning frameworks.

It is worth noting that our bidirectional decomposition is scalable and not limited to the middle plane. For example, we can manually control the virtual plane closer to the reference or target plane, allocating more distortion to the less salient view and thus producing a more natural appearance in visual perception.

\subsubsection{Loss Function}
The total loss function of our spatial model consists of an alignment loss and a shape-preserving loss as follows:
\begin{equation}
    \mathcal{L}^{S} = \mathcal{L}^{S}_{align} + \lambda \mathcal{L}^{S}_{shape}.
    \label{eq_spatial_warp}
 \end{equation}
 
The alignment component leverages photometric errors to encourage the predicted homography, and TPS warps can align the input images as:
\begin{equation}
\begin{aligned}
    \mathcal{L}_{align}^{S} =  \omega &\Vert{\small \mathcal{W}(I^t_{ref}, H_{ref}(t))}-{\small \mathcal{W}(I^t_{tgt}, H_{tgt}(t))}\Vert_1 \cdot{\small OP_h}  \\
     + & \Vert{\small \mathcal{W}(I^t_{ref}, m_{ref}^{S}(t))}-{\small \mathcal{W}(I^t_{tgt}. m_{tgt}^{S}(t))}\Vert_1 \cdot{\small OP_{tps}}.\\
    \end{aligned}
 \end{equation}
Here, 
\begin{equation}
OP_h = {\small \mathcal{W}(\mathds{1}, H_{ref}(t))}\cdot{\small \mathcal{W}(\mathds{1}, H_{tgt}(t))},
 \end{equation}
and
\begin{equation}
OP_{tps} = {\small \mathcal{W}(\mathds{1}, m_{ref}^{S}(t))}\cdot{\small \mathcal{W}(\mathds{1}, m_{tgt}^{S}(t))}.
\label{eq:op_tps}
 \end{equation}
$\mathcal{W}(\cdot,\cdot)$ is the warping process, and $\mathds{1}$ is an all-one matrix with the same resolution as the input image. $OP_h$ and $OP_{tps}$ denote the overlapping regions in the global and local parts, respectively. 
We adopt the mesh term \cite{nie2022deep} (named $\mathcal{L}_{distortion}(\cdot)$ in our formulations) to calculate our shape-preserving loss, which can be written as:
\begin{equation}
    \mathcal{L}_{shape}^{S} =  \mathcal{L}_{distortion}(m_{ref}^{S}(t)) + \mathcal{L}_{distortion}(m_{tgt}^{S}(t)).
    \label{eq:spatial_distortion}
 \end{equation}

\subsection{Temporal Warp}

\subsubsection{Network}
\textit{StabStitch} warps target frames to align with reference frames, so we only need to obtain temporal warps between consecutive target frames. In contrast, \textit{StabStitch++} simultaneously warp the reference and target frames, which requires the temporal warps from consecutive reference and target frames. To this end, we simplify our temporal warp model to reduce the elapsed time. The network is illustrated in Fig. \ref{fig:network} (b), where we take $(t\!-\!1)$-th and $t$-th frame ($I^{t\!-\!1}$, $I^t$) as input and output the temporal warps $m^T(t)$. We also use the control point motions of TPS transformation to represent the temporal warps, which hold the same resolution of control points as that of the spatial model. Compared with the spatial warp model, our temporal model removes the global homography estimation and only predicts a unidirectional warp to align $I^t$ with $I^{t\!-\!1}$. For brevity, we omit the subscript of reference or target frames ($ref/tgt$) in this section.

\subsubsection{Loss Function}
Similar to the spatial warp model, the loss function of our temporal model also consists of two components as follows:
\begin{equation}
    \mathcal{L}^{T} = \mathcal{L}^{T}_{align} + \lambda \mathcal{L}^{T}_{shape}.
 \end{equation}
Here,
\begin{equation}
    \mathcal{L}_{align}^{T}  = \Vert{\small I^{t\!-\!1}-{\small \mathcal{W}(I^{t}, m^T(t))}\Vert_1 \cdot{\small \mathcal{W}(\mathds{1}, m^T(t))}},
 \end{equation}
and,
\begin{equation}
    \mathcal{L}_{shape}^{T} =  \mathcal{L}_{distortion}(m^{T}(t)).
    \label{eq:temporal_distortion}
 \end{equation}

\subsection{Stitching Trajectory Generation}

\subsubsection{Camera Trajectory}
Camera trajectory is widely used in video stabilization and can be defined as a chain of relative motions, such as Euclidean transformations \cite{liu2012video}, homography transformations \cite{liu2013bundled}, etc. Representing the transformation of the initial frame as an identity matrix $F(1)$, the camera trajectories are written as:
\begin{equation}
    C(t)=F(1)F(2)\cdot \cdot \cdot F(t),
    \label{eq_paths_homo}
 \end{equation}
where $F(t)$ is the relative transformation from the $t$-th frame to the $(t\!-\!1)$-th frame. Considering that our temporal model directly predicts 2D motions of each control point, we adopt the motion representation of vertex files like MeshFlow \cite{liu2016meshflow}. Concretely, we chain the temporal motions of control points as camera trajectories for a straightforward representation:
\begin{equation}
    C(t)=m^T(1)+m^T(2)+\cdot \cdot \cdot +m^T(t),
    \label{eq_paths_meshflow}
 \end{equation}
where $m^T(1)$ is set to all-zero matrix. Note each control point in $m^T(t)$ is anchored at every vertex in a rigid mesh. 

\subsubsection{Stitching Trajectory}
Video stabilization leverages the chain of temporal motions as camera paths, whereas in our video stitching system, it throws a question of how to represent the stitching paths of a warped video. We delve into this question by combining the spatial and temporal warp models. 
Here, we only derive the formulation of stitching paths for the reference view, and the trajectories for the target view can be obtained similarly. Particularly, we first reach the spatial/temporal motions $m_{ref}^T(t)$, $m_{ref}^S(t\!-\!1)$, $m_{ref}^S(t)\in \mathbb{R}^{2\times (U+1) \times (V+1)}$ and their corresponding meshes $M_{ref}^T(t)$, $M_{ref}^S(t\!-\!1)$, $M_{ref}^S(t)\in \mathbb{R}^{2\times (U+1) \times (V+1)}$ as follows:
\begin{equation}
  \label{eq:stitchmeshflow1}
  \begin{matrix}
    \begin{aligned}
        &\begin{cases}
        m_{ref}^{T}(t) =  TNet(I_{ref}^{t\!-\!1}, I_{ref}^{t}),\\
      m_{ref}^{S}(t\!-\!1), m_{tgt}^{S}(t\!-\!1) = SNet(I_{ref}^{t\!-\!1}, I_{tgt}^{t\!-\!1}), \\
      m_{ref}^{S}(t), m_{tgt}^{S}(t) = SNet(I_{ref}^{t}, I_{tgt}^{t}),
      \end{cases}
      \quad \Rightarrow
      \\
      &\begin{cases}
      M_{ref}^{T}(t) = M^{Rig}+m_{ref}^{T}(t),\\ 
      M_{ref}^{S}(t\!-\!1) = M^{Rig}+m_{ref}^{S}(t\!-\!1),\\
      M_{ref}^{S}(t) = M^{Rig}+m_{ref}^{S}(t),
      \end{cases}
    \end{aligned}\\ 
  \end{matrix}
\end{equation}
where $SNet$/$TNet(\cdot,\cdot)$ represents the spatial/temporal warp model, and $M^{Rig}\in \mathbb{R}^{2\times (U+1) \times (V+1)}$ is defined as the pixel positions of control points in a rigid mesh. 

Then, we need to derive the stitching motions of warped videos from the spatial/temporal meshes. To align the $t$-th frame with the $(t\!-\!1)$-th frame in the warped video, the temporal meshes from the $t$-th frame to the $(t\!-\!1)$-th frame $M_{ref}^{T}(t)$ should also undergo the same transformations as the spatial warps of the $(t\!-\!1)$-th frame $M_{ref}^{S}(t\!-\!1)$. Assuming $\mathcal{T}(\cdot)$ is the TPS transformation, the desired stitching motions could be represented as the difference between the desired meshes and the actual spatial meshes $M_{ref}^{S}(t)$: 
\begin{equation}
\begin{aligned}
    s_{_{ref}}(t)=&\mathcal{T}_{M^{Rig}\to M_{ref}^{S}(t\!-\!1)}(M_{ref}^{T}(t)) - M_{ref}^{S}(t),\\
    \end{aligned}
    \label{eq:stitchmeshflow2}
 \end{equation}
Finally, we attain the stitching trajectories by chaining the relative stitching motions between consecutive warped frames as follows:
 \begin{equation}
 \begin{aligned}
    S_{ref}(t)=&s_{_{ref}}(1)+s_{_{ref}}(2)+\cdot \cdot \cdot +s_{_{ref}}(t), \\
    \end{aligned}
    \label{eq:stitchmeshflow3}
 \end{equation}
where we define $s_{_{ref}}(1)$ is an all-zero matrix.

Temporal shakes typically arise from discontinuities in sequential trajectories. As evidenced by Eq. \ref{eq:stitchmeshflow2}, we can observe: (1) When the spatial warp degenerates to a constant or becomes non-existent, the stitching trajectories reduce to conventional camera trajectories~\cite{liu2016meshflow} used in video stabilization. At this time, the shakes are only from the source videos. (2) When the spatial warps at different timestamps are different or varying irregularly, even if the source videos are stable, temporal shakes will be produced from warping.

\subsection{Warp Smoothing}
To get a stable warped video, we need to smooth the stitching trajectories and preserve the natural shapes after warping. 

\vspace{0.2cm}
\subsubsection{Model Achitecure}
In this stage, a warp smoothing model is designed to achieve the above goals. As depicted in Fig. \ref{fig:framework}, it takes sequences of ($N$ frames) stitching paths ($S_{ref}$, $S_{tgt}$) and spatial meshes ($M_{ref}^S$, $M_{tgt}^S$) as input, and then outputs the smoothing increments ($\Delta_{ref}$, $\Delta_{tgt}$) as described in the following equation:
 \begin{equation}
    \Delta_{ref}, \Delta_{tgt} = SmoothNet(S_{ref}, S_{tgt}, M_{ref}^S, M_{tgt}^S),
    \label{eq_smoothnet1}
 \end{equation}
where $S_{ref}$,$M_{ref}^S$,$\Delta_{ref}\in \mathbb{R}^{2\times N \times (U+1) \times (V+1)}$.

The network architecture is shown in Fig. \ref{fig:network}(d). This model first embeds $S_{ref}$, $S_{tgt}$,$M_{ref}^S$, $M_{tgt}^S$ into 32 channels through separate linear projections. Then, these embeddings are concatenated and fed into three 3D convolutional layers to model the spatiotemporal dependencies. Finally, we reproject the hidden results back into 4 channels to get $\Delta_{ref}$ and $\Delta_{tgt}$. The network architecture is designed so compact that it can accomplish efficient smoothing inference. In addition, this simple architecture better highlights the effectiveness of the proposed unsupervised learning scheme.

With the smoothing increment $\Delta_{ref}$, we define the smooth stitching paths for the reference view as:
 \begin{equation}
    \hat{S}_{ref} = S_{ref} + \Delta_{ref}.
    \label{eq_smoothnet2}
 \end{equation}
Then we can expand Eq. \ref{eq_smoothnet2} based on Eq. \ref{eq:stitchmeshflow3} and Eq. \ref{eq:stitchmeshflow2}, and obtain:
\begin{equation}
  \label{eq_smoothnet3}
  \begin{matrix}
    \begin{aligned}
      \hat{S}_{ref}(t) &=  S_{ref}(t\!-\!1) + s_{_{ref}}(t) + \Delta_{ref}(t) =  S_{ref}(t\!-\!1) +\\
      &\mathcal{T}_{M^{Rig}\to M_{ref}^{S}(t\!-\!1)}(M_{ref}^{T}(t)) - \underbrace{(M_{ref}^{S}(t) - \Delta_{ref}(t))}_{\text{Smooth spatial mesh}}.
    \end{aligned}\\ 
  \end{matrix}
\end{equation}
In this case, the last term in Eq. \ref{eq_smoothnet3} can be regarded as the smooth spatial mesh $\hat{M}_{ref}^{S}(t)$. Hence, the sequences of smooth spatial meshes for the reference view are written as:
\begin{equation}
    \hat{M}_{ref}^{S} = M_{ref}^{S} - \Delta_{ref}.
    \label{eq_smoothnet4}
 \end{equation}
As for the target view, the corresponding smooth stitching paths and meshes can be calculated in a similar way.

\vspace{0.2cm}
\subsubsection{Objective Function}
Built on the warp smoothing model as described in Eq. \ref{eq_smoothnet1}, we define a comprehensive objective function as the balance of different unsupervised constraints:
\begin{equation}
    \mathcal{L} =  \mathcal{L}_{data} + \omega_{1} \mathcal{L}_{smooth} + \omega_{2} \mathcal{L}_{shape} + \omega_{3} \mathcal{L}_{trajectory}.
    \label{eq_smoothgoal1}
\end{equation}

\vspace{0.1cm}
\noindent\textbf{\colorbox[rgb]{0.93,0.93,0.93}{Data Term:}}
The data term encourages the final paths (\textit{i.e.}, $\hat{S}_{ref}$, $\hat{S}_{tgt}$) to be close to the original paths. This constraint alone does not contribute to stabilization. The stabilizing effect of \textit{StabStitch++} is realized in conjunction with the data term and the subsequent smoothness term. It is formulated as follows:
\begin{equation}
    \mathcal{L}_{data} = \Vert\hat{S}_{ref}-S_{ref}\Vert_2 + \Vert\hat{S}_{tgt}-S_{tgt}\Vert_2.
    \label{eq_smoothgoal2}
 \end{equation}

\noindent\textbf{\colorbox[rgb]{0.93,0.93,0.93}{Smoothness Term:}}
In a smooth path, each motion should not contain sudden large-angle rotations, and the amplitude of translations should be as consistent as possible. To this end, we encourage the trajectory position at a certain moment to be located at the midpoint between its positions in the preceding and succeeding moments, which implicitly satisfies the above two requirements. It is defined as:
 \begin{equation}
  \label{eq_smoothgoal3}
  \begin{matrix}
    \begin{aligned}
      \mathcal{L}_{smooth} =  &\sum_{i=1}^{\bar{m}\!-\!1} \alpha_{i}\Vert \hat{S}_{ref}(\bar{m}\!+\!i) + \hat{S}_{ref}(\bar{m}\!-\!i) - 2\hat{S}_{ref}(\bar{m})\Vert_2 \\
      +&\sum_{i=1}^{\bar{m}\!-\!1} \alpha_{i}\Vert \hat{S}_{tgt}(\bar{m}\!+\!i) + \hat{S}_{tgt}(\bar{m}\!-\!i) - 2\hat{S}_{tgt}(\bar{m})\Vert_2,
    \end{aligned}
  \end{matrix}
\end{equation}
where $\bar{m}$ is the middle index of $N$ ($N$ is required to be an odd number) and $\alpha_{i}$ is a constant between 0 and 1 to impose smoothing constraints of varying significance on trajectories at different temporal intervals.

\vspace{0.1cm}
\noindent\textbf{\colorbox[rgb]{0.93,0.93,0.93}{Shape Preserving Term:}}
The warping shakes can be effectively removed under the balance of data and smoothness terms.
However, the trajectory of each control point is optimized individually. Actually, each warped video has $(U+1)\times(V+1)$ control points, which means there are $(U+1)\times(V+1)$ independently optimized trajectories. When these trajectories are changed inconsistently, significant distortions will be produced. To remove the distortions and encourage different paths to share similar changes, we introduce a shape-preserving term as:
 \begin{equation}
  \label{eq_smoothgoal4}
  \begin{matrix}
    \begin{aligned}
      \mathcal{L}_{shape} = &\frac{1}{N} \sum_{t=1}^{N} \mathcal{L}_{distortion}(\hat{M}_{ref}^{S}(t)-M^{Rig})+\\
      &\frac{1}{N} \sum_{t=1}^{N} \mathcal{L}_{distortion}(\hat{M}_{tgt}^{S}(t)-M^{Rig}),
    \end{aligned}
  \end{matrix}
\end{equation}
where $\mathcal{L}_{distortion}(\cdot)$ takes mesh motions as input and calculates the mesh distortion as used in Eq. \ref{eq:spatial_distortion} and Eq. \ref{eq:temporal_distortion}.

\vspace{0.1cm}
\noindent\textbf{\colorbox[rgb]{0.93,0.93,0.93}{Trajectory Consistency Term:}} The shape-preserving term encourages the trajectories of all control points in a warped video change consistently. However, there might be inconsistent trajectories between the reference warped video and the target one. To address this potential issue, we design a trajectory consistency term. Concretely, we only constrain the trajectory consistency in overlapping areas between different views. However, under the joint effect of the shape-preserving term, the trajectory consistency of overlapping areas can be extended to the complete views.

Besides, these trajectories are anchored to sparse control points, and the control points from different views in the overlapping area hardly overlap after warping. Therefore, these sparse trajectories from different views are not located in the same position. To this end, we resample the sparse trajectories to pixel-level dense trajectories and then apply the following constraints:
\begin{equation}
\begin{aligned}
    \mathcal{L}_{trajectory}  = &\Vert \mathcal{W}(\uparrow\!(\hat{S}_{ref}(t)), \hat{M}_{ref}^S(t)\!-\!M^{Rig})- \\
    &\mathcal{W}(\uparrow\!(\hat{S}_{tgt}(t)), \hat{M}_{tgt}^S(t)\!-\!M^{Rig})\Vert_1 \cdot \hat{OP}_{tps},
    \end{aligned}
 \end{equation}
where $\uparrow\!(\cdot)$ is the resampling operation and $\hat{OP}_{tps}$ denotes the overlapping regions that can be obtained following Eq. \ref{eq:op_tps}.

%% file: sections/Online.tex

\begin{figure*}[t]
	\centering
	\includegraphics[width=0.99\linewidth]{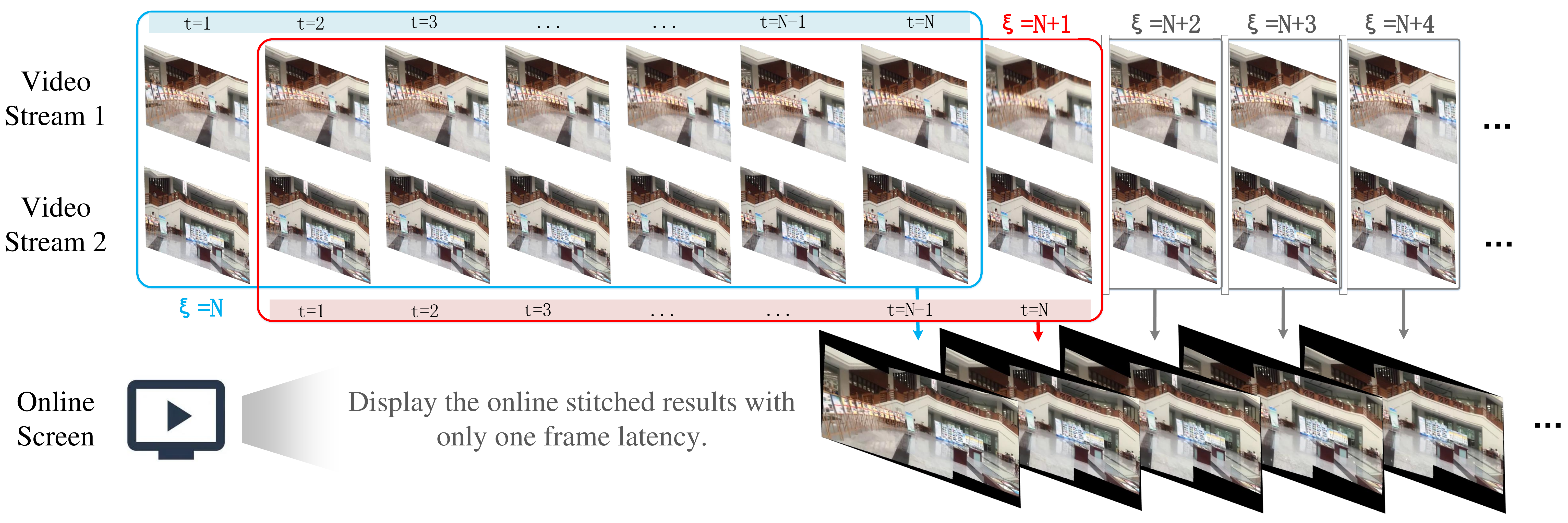}
 \vspace{-0.35cm}
	\caption{The online stitching mode. We define a sliding window to process a short sequence and display the last frame on the online screen.}
	\label{fig:online}
  \vspace{-0.3cm}
\end{figure*}

\section{Online Stitching}
\label{sec:online}

Existing video stitching methods \cite{nie2017dynamic}\cite{su2016video}\cite{guo2016joint}\cite{lin2016seamless}\cite{jiang2015video} are offline solutions, which smooth the trajectories after the videos are completely captured. 
Besides, the final result typically takes a long time (much longer than the duration of the video).
Unlike them, \textit{StabStitch++} is an online video stitching solution that does not require the subsequent frames after the current frame and stitches videos in real time.

\subsection{Online Smoothing}
To realize online inference, we define a fixed-length sliding window ($N$ frames) to cover previous $N\!-\!1$ frames and the current frame, as shown in Fig. \ref{fig:online}. Then, the local stitching trajectory inside this window is extracted and smoothed according to Sec. \ref{sec:Methodology}. Next, warped current frames are rendered using the smooth spatial meshes $\hat{M}_{ref}^S(N)$, $\hat{M}_{tgt}^S(N)$. Finally, we blend them to get the stable stitched result (of the current frame) and display it when the next frame arrives. With this mode and efficient architectures, \textit{StabStitch++} can achieve minimal latency with only one frame. To support online stitching, we redesign the objective function of the warp smoothing model (Eq. \ref{eq_smoothgoal1}) with two online terms as follows:

\begin{equation}
\begin{aligned}
        \mathcal{L} =  &\mathcal{L}_{data} + \omega_{1} \mathcal{L}_{smooth} + \omega_{2} \mathcal{L}_{shape} +\\
        &\omega_{3} \mathcal{L}_{trajectory} + \omega_{4} \mathcal{L}_{online} + \omega_{5} \mathcal{L}_{align}.
\end{aligned}
\end{equation}

\vspace{0.1cm}
\noindent\textbf{\colorbox[rgb]{0.93,0.93,0.93}{Online Collaboration Term:}}
This first term is an online collaboration term. Compared with the previous offline mode, the online mode could introduce a new issue, wherein the smoothed trajectories in different sliding windows (with partial overlapping sequences) may be inconsistent.
It could produce subtle jitters if we chain the last frame of all sliding windows to form a complete stitched video. Therefore, we design this online collaboration term to address the above issue as:
 \begin{equation}
  \label{eq_online1}
  \begin{matrix}
    \begin{aligned}
      \mathcal{L}_{online} = &\frac{1}{N-1} \sum_{t=2}^{N} \Vert \hat{S}_{ref}^{(\xi)}(t) - \hat{S}_{ref}^{(\xi+1)}(t-1) \Vert_2 + \\
      &\frac{1}{N-1} \sum_{t=2}^{N} \Vert \hat{S}_{tgt}^{(\xi)}(t) - \hat{S}_{tgt}^{(\xi+1)}(t-1) \Vert_2 ,
    \end{aligned} 
  \end{matrix}
\end{equation}
where $\xi$ is the absolute time ranging from $N$ to the last frame of the videos and also implies the index of sliding windows. In contrast, $t$ can be regarded as the relative time in a certain sliding window ranging from $1$ to $N$.

\vspace{0.1cm}
\noindent\textbf{\colorbox[rgb]{0.93,0.93,0.93}{Online Alignment Term:}}
In our previous conference version \cite{nie2024eliminating}, \textit{StabStitch} first carried out pre-alignment and then struggled to preserve the alignment performance while smoothing the stitching trajectories. Different from that, \textit{StabStitch++} makes no compromise and simultaneously optimizes both of them in the warp smoothing model. Although it only takes the low-resolution trajectories and meshes as input to ensure inference efficiency, we leverage high-resolution input images to impose implicit guidance on the deformed meshes, which is proved to be effective in our experiments. We call it an online alignment term and formulate it as follows:
\begin{equation}
\begin{aligned}
    \mathcal{L}_{align}  = &\Vert \mathcal{W}(I_{ref}^{N}, \hat{M}_{ref}^{S}(N)\!-\!M^{Rig})- \\
    &\mathcal{W}(I_{tgt}^{N}, \hat{M}_{tgt}^{S}(N)\!-\!M^{Rig})\Vert_1 \cdot \hat{OP}_{tps}.
    \end{aligned}
 \end{equation}
It is worth noting that we only apply this alignment constraint to the last frame of a sliding window. The effect is comparable to imposing this constraint on all frames, but it reduces massive redundant gradients in the training process.


\subsection{Offline and Online Inference}
Offline smoothing takes the whole trajectories as input, outputs the optimized whole trajectories, and then renders all the video frames. It conducts smoothing after receiving the whole input videos and can be regarded as a special online case in which the sliding window covers all video frames. By contrast, online smoothing takes local trajectories as input, outputs the smoothed local trajectories, and then only renders the last frame in the sliding window. The future frames beyond the current frames are unseen to the online smoothing process. It is more challenging than offline inference because the future frames beyond the current frames are not available in the online smoothing process, and this mode allows real-time playback of the stitched video while capturing input videos.

%% file: sections/Experiment.tex
\section{Experiment}
\label{sec:Experiment}
In this section, we first introduce the proposed dataset and other datasets used in our paper, and then describe the experimental details and metrics. Then we carry out extensive comparative experiments with SoTA solutions. Subsequently, the ablation studies are depicted to show our effectiveness. 

\begin{figure*}[t]
	\centering
	\includegraphics[width=0.94\linewidth]{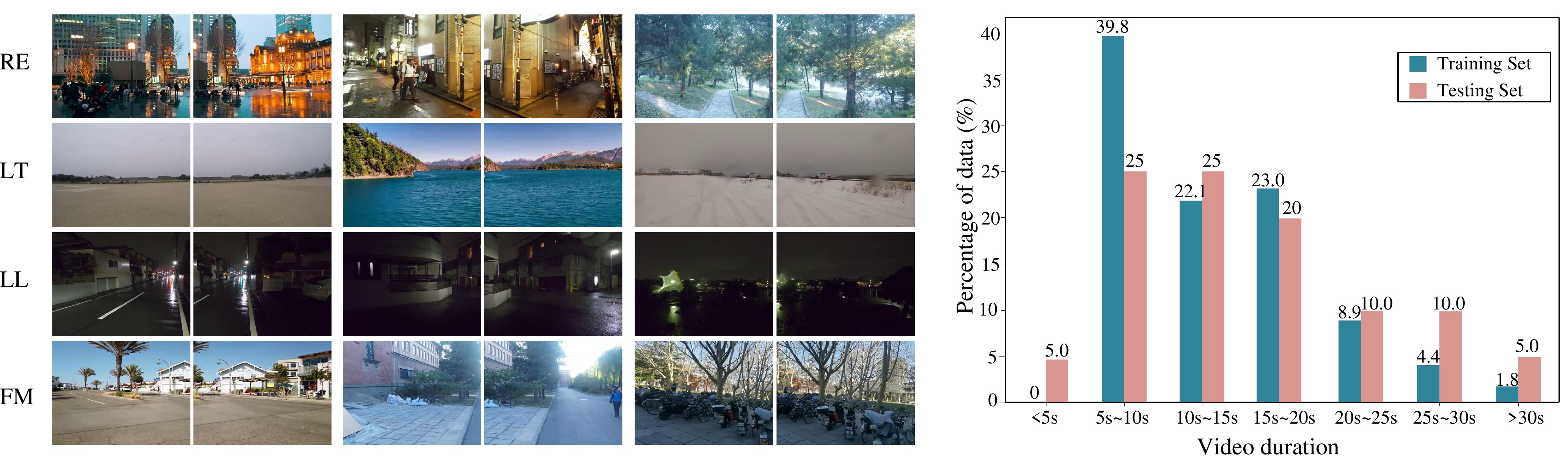}
 \vspace{-0.25cm}
	\caption{The proposed \textit{\textbf{StabStitch-D}} dataset. Left: several video examples from diverse scenes. Right: the distribution of video duration time.}
	\label{fig:dataset}
  \vspace{-0.3cm}
\end{figure*}

\subsection{Dataset Preperation}
\paragraph{StabStitch-D} 
Considering the lack of dedicated datasets for video stitching, we establish \textit{\textbf{StabStitch-D}}, a comprehensive dataset for training and evaluation. Our dataset comprises over 100 video pairs, consisting of over 100,000 images, with each video lasting from approximately 5 seconds to 35 seconds. To holistically reveal the performance of video stitching methods in various scenarios, we categorize videos into four classes based on their content, including regular (RE), low-texture (LT), low-light (LL), and fast-moving (FM) scenes. In the testing split, 20 video pairs are divided for testing, with 5 videos in each category. Fig. \ref{fig:dataset}(left) illustrates some examples for each category, where FM is the most challenging case with fast irregular camera movements (rotation or translation). The distribution statistics of video duration are demonstrated in Fig. \ref{fig:dataset}(right).
Each video's resolution is resized into $360\times 480$ for efficient training, and arbitrary resolutions are supported in the testing phase.

\paragraph{Traditional Dataset} 
The videos in \textit{\textbf{StabStitch-D}} are typically stable due to the advancement of video stabilization in software and hardware. Considering that, we also collect some unstable videos from traditional video stitching datasets \cite{guo2016joint, nie2017dynamic, su2016video, lin2016seamless} to further validate our performance. These videos are captured by different smartphones and drones with a wide range of resolutions such as $540\times 960$, $720\times 1280$, $1080\times1920$, etc. There are a total of 31 unstable video pairs, each of which lasts for more than 10 seconds. These videos are challenging cases that are extremely shaky, low overlapped, or contain dynamic objects.

\subsection{Implementation Detail}
\label{details}
\paragraph{Detail} 
We implement the whole framework in PyTorch with one RTX 4090Ti GPU. The spatial warp, temporal warp, and warp smoothing models are trained separately, with the epoch number set to 40, 100, and 50. The parameter sizes of each model are 42.56MB, 24.51MB, and 8.45MB, respectively. Initially, we train the first two warp models (spatial and temporal) and adopt the pre-trained models to predict spatial and temporal meshes (\textit{i.e.}, $M_{ref}^S(t)$, $M_{tgt}^S(t)$, $M_{ref}^T(t)$, $M_{tgt}^T(t)$) for each frame. These predicted meshes are then leveraged to generate stitching trajectories, which further serve as the input of the warp smoothing model. We finally train the warp smoothing models with multiple objective terms.
$\lambda$ and $\omega$ are defined as 10 and 3. The weights for data, smoothness, shape preservation, online collaboration, trajectory consistency, and online alignment terms are set to 1, 50, 10, 0.1, 10, and 1000. $\alpha_1$, $\alpha_2$, and $\alpha_3$ are set to 0.9, 0.3, and 0.1. The control point resolution and sliding window length are empirically set to $(6+1)\times(8+1)$ and 7. 

\paragraph{Augmentation} 
When training the warp smoothing model, we carry out data augmentation by randomly selecting $N=7$ frames as the sliding window from a buffer of 12 frames, which could allow more diverse stitching trajectories.

\subsection{Metric}
To evaluate the proposed solution quantitatively, we suggest three metrics: alignment, distortion, and stability. 

\vspace{0.1cm}
\noindent\textbf{Alignment Score}: Following the criterion of UDIS \cite{nie2021unsupervised} and UDIS++ \cite{nie2023parallax}, we also adopt PSNR and SSIM of the overlapping regions to evaluate the alignment performance. We average the scores in all video frames.

\vspace{0.1cm}
\noindent\textbf{Distortion Score}: The final warps in the online stitching mode can be described as a series of meshes: $\hat{M}^{S(N)}(N)$, $\hat{M}^{S(N+1)}(N)$, $\cdot\cdot\cdot$, $\hat{M}^{S(\xi)}(N)$, $\cdot\cdot\cdot$. Then we adopt $\mathcal{L}_{shape}(\cdot)$ to measure the distortion magnitude. Because any distortion in a single frame will destroy the perfection of the whole video, we choose the mean value of the maximum distortion values of each video as the distortion score. 

\vspace{0.1cm}
\noindent\textbf{Stability Score}: The smoothed trajectories in the online stitching mode can also be described as a series of positions: $\hat{S}^{(N)}(N)$, $\hat{S}^{(N+1)}(N)$, $\cdot\cdot\cdot$, $\hat{S}^{(\xi)}(N)$, $\cdot\cdot\cdot$. Then we adopt $\mathcal{L}_{smooth}(\cdot)$ to measure the stability. The stability score is the mean value of the average smoothness loss of each video.

\vspace{0.1cm}
To intuitively compare with our conference version (\textit{i.e.}, \textit{StabStitch} \cite{nie2024eliminating}), we only calculate the metrics of target frames for distortion and stability scores.

\subsection{Compared with Current Solutions} 
\label{results}
We compare our method with image and video stitching solutions, respectively.

\subsubsection{Compared with Image Stitching}
Two representative SoTA image stitching methods are selected to compare with our solution: LPC \cite{jia2021leveraging} (traditional method), and UDIS++ \cite{nie2023parallax} (learning-based method).
The quantitative comparison results are illustrated in Tab. \ref{quantitative results}, where `$\cdot/\cdot$' indicates the PSNR/SSIM values. `-' implies the approach fails in this category (\textit{e.g.}, program crash or extremely severe distortion). The results show that our solution achieves better alignment performance than the current SoTA image stitching methods. 


\setlength{\tabcolsep}{4pt}
\begin{table*}[t]
	\begin{center}
		\renewcommand{\arraystretch}{1.1}
		\caption{Quantitative comparisons with image stitching methods on  \textit{StabStitch-D} dataset. * indicates the model is re-trained on the proposed dataset.}
		\label{quantitative results}
    \scalebox{1}{
		\begin{tabular}{ccccccc}
			\hline
			 &Method & Regular & Low-Light & Low-Texture & Fast-Moving & Average  \\
			\cline{2-7}
            1 & LCP \cite{jia2021leveraging}  & 24.22/0.812 & - & - & 23.88/0.813 & - \\
			2 & UDIS++ \cite{nie2023parallax}  & 23.19/0.785 & 31.09/0.936 & 29.98/0.906 & 21.56/0.756 & 27.19/0.859 \\
            3 & UDIS++ * \cite{nie2023parallax}  & 24.63/0.829 & 34.26/0.957 & 32.81/0.920 & 24.78/0.819 & 29.78/0.891 \\
            4 & StabStitch++  & $\mathbf{25.51}$/$\mathbf{0.837}$ & $\mathbf{35.10}$/$\mathbf{0.958}$ & $\mathbf{34.23}$/$\mathbf{0.928}$ & $\mathbf{25.64}$/$\mathbf{0.830}$ & $\mathbf{30.80}$/ $\mathbf{0.897}$ \\
			\hline
		\end{tabular}
  }
	\end{center}
 \vspace{-0.5cm}
\end{table*}



\begin{table}[!t]
  \centering
  \caption{ User study of the preference on  \textit{StabStitch-D} dataset. We exclude the failure cases of Nie \textit{et\ al.} \cite{nie2017dynamic} for fairness.}
  \vspace{-0.2cm}
  \renewcommand{\arraystretch}{1.2}
  \begin{tabular}{ccc}
   \hline
    StabStitch++ &  Nie \textit{et\ al.} \cite{nie2017dynamic} & No preference \\
   \hline
  34.38\% & 3.91\%  & 61.71\% \\
      \hline
   \end{tabular}
   \vspace{-0.4cm}
   \label{table:user}
\end{table}

\setlength{\tabcolsep}{4pt}
\begin{table}[t]
	\begin{center}
		\renewcommand{\arraystretch}{1.1}
		\caption{Quantitative comparisons with video stitching methods on  \textit{StabStitch-D} dataset.}
        \vspace{-0.2cm}
		\label{video_compare}
		\begin{tabular}{ccccc}
			\hline
			 & Method & Alignment $\uparrow$ & Stability $\downarrow$ & Distortion $\downarrow$  \\
			\cline{2-5}
            1  & StabStitch \cite{nie2024eliminating}& 29.89/0.890 & 48.74 & 0.674  \\
            2  & StabStitch++ & $\mathbf{30.88}$/$\mathbf{0.898}$ & $\mathbf{41.70}$ & $\mathbf{0.371}$  \\
			\hline
		\end{tabular} 
	\end{center}
  \vspace{-0.5cm}
\end{table}

\subsubsection{Compared with Video Stitching} 
\begin{figure*}[t]
	\centering
    
	\includegraphics[width=0.99\linewidth]{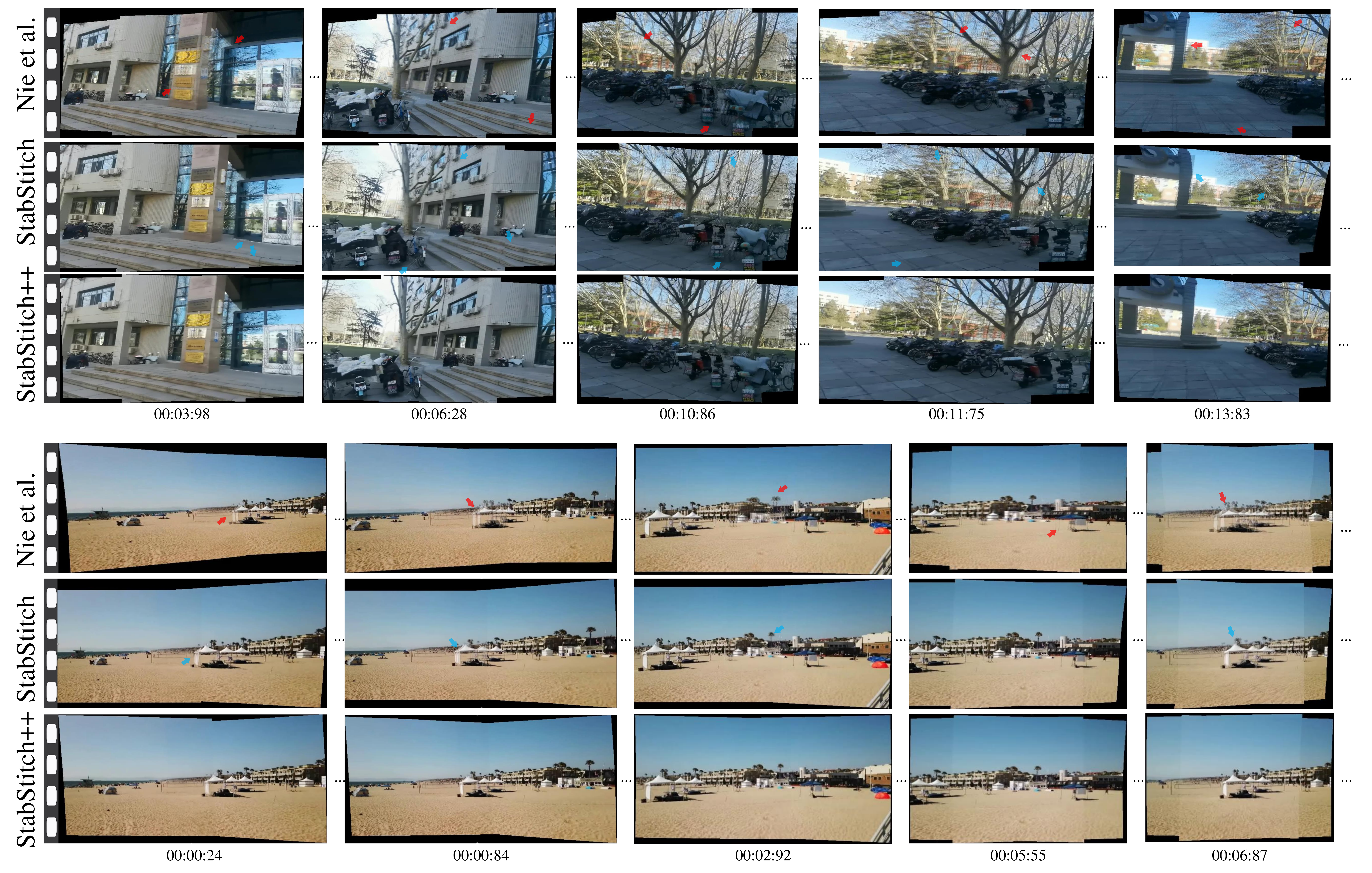}
 \vspace{-0.25cm}
	\caption{Qualitative comparison with Nie \textit{et\ al.}'s video stitching \cite{nie2017dynamic} and  \textit{StabStitch} \cite{nie2024eliminating} on the  \textit{StabStitch-D} dataset. The arrows indicate artifacts or distortions, and the numbers below the images exhibit the time at which the frame appears in the video. Please zoom in for the best view.}
	\label{fig:comparison}
\end{figure*}

Next, we compare our method with video stitching methods, including Nie \textit{et\ al.}'s video stitching solution \cite{nie2017dynamic} (traditional method) and  \textit{StabStitch} \cite{nie2024eliminating} (learning-based method). To our knowledge, they are the latest and best video stitching methods for hand-held cameras. 
For  \textit{StabStitch} \cite{nie2024eliminating}, we conduct quantitative and qualitative comparisons with it. As illustrated in Tab. \ref{video_compare} and Fig. \ref{fig:comparison}, the proposed  \textit{StabStitch++} gains comprehensive performance improvements.
As for Nie \textit{et\ al.}'s video stitching solution \cite{nie2017dynamic}, it tends to fail in specific challenging cases, such as low texture or low light. Therefore, we replace the quantitative experiments with user studies and demonstrate the qualitative results in Tab. \ref{table:user} and Fig. \ref{fig:comparison}.

\begin{figure*}[t]
	\centering
    
	\includegraphics[width=0.96\linewidth]{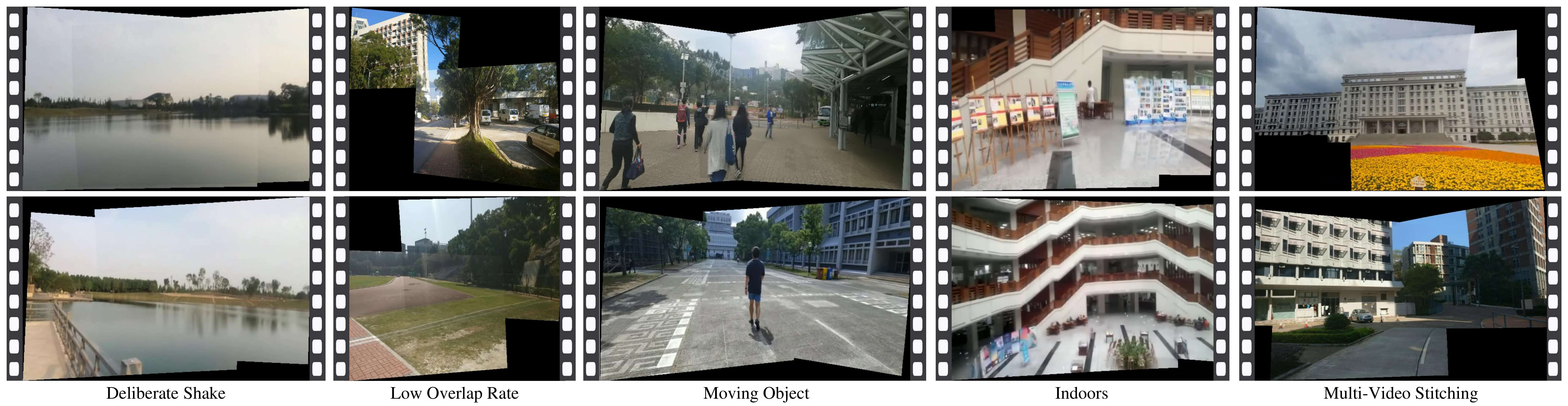}
 \vspace{-0.25cm}
	\caption{Our video stitching results on traditional datasets. Refer to the supplementary video for more details.}
	\label{fig:tra_data}
\end{figure*}

\vspace{0.1cm}
In particular, in the testing set of  \textit{StabStitch-D} dataset \cite{nie2024eliminating} (20 video cases in total), Nie \textit{et\ al.} \cite{nie2017dynamic} fail in 10 video cases because of program crashes. Hence, we exclude these failure cases and conduct the user study only on the successful cases. For a stitched video, different methods may perform differently at different times. To this end, we segment each complete stitched video into one-second clips (we omit the last clip of a stitched video that is shorter than one second in practice), resulting in 128 clips in total. Then, we invite 20 participants, including 10 researchers/students with computer vision backgrounds and 10 volunteers outside this community. In each test session, two clips from different methods are presented randomly, and every volunteer is required to indicate their overall preference for alignment, distortion, and stability. We average the preference rates and exhibit the results in Tab. \ref{table:user}, where our results are clearly preferred. 


\begin{table*}[!t]
  \centering
  \caption{A comprehensive analysis of inference speed.}
  \vspace{-0.2cm}
  \renewcommand{\arraystretch}{1.2}
  \begin{tabular}{ccccccccc}
   \hline
    & Component & SNet &  TNet & Trajectory generation & SmoothNet & Warping & Blending & Total\\
   \hline
   1& StabStitch \cite{nie2024eliminating} &11.5ms &  10ms &  1.1ms & 1ms & 4.4ms & 0.2ms& 28.2ms \\
   2 & StabStitch++ &16.1ms & 6.9ms  & 2.2ms  & 1.4ms & 8.5ms & 0.2ms & 35.3ms \\
      \hline
   \end{tabular}
   \vspace{-0.3cm}
   \label{table:speed}
\end{table*}

\subsection{Analysis}
\subsubsection{More Result}
We further evaluate our solution on traditional datasets, where the videos are collected from previous video stitching works \cite{guo2016joint, nie2017dynamic, su2016video, lin2016seamless}. Due to the limitations of previous video stabilization technologies, almost all input videos are shaky, and some are even intentionally shaky. Even though this situation does not conform to the current technology state, our method can still stably stitch them together, especially in various challenging scenes such as deliberate shakes, low overlap rates, dynamic objects, etc. The results are simply exhibited in Fig. \ref{fig:tra_data}. Please refer to the supplementary video for more details.

\subsubsection{Inference Speed}
A comprehensive analysis of the inference speed is provided in Tab. \ref{table:speed} with one RTX 4090Ti GPU, where `Blending' represents the average blending. 
We conduct the experiment on the third case of category `RE' in Fig. \ref{fig:dataset}. As shown in Tab. \ref{table:speed}, \textit{StabStitch} \cite{nie2024eliminating} only takes about 28.2ms to stitch one frame, yielding a real-time online video stitching system. When stitching a video pair with higher resolution, only the time for warping and blending steps will slightly increase. In contrast, Nie \textit{et\ al.}'s solution \cite{nie2017dynamic} takes over 40 minutes to get such a 26-second stitched video with an Intel i7-9750H 2.60GHz CPU, making it impractical to be applied to online stitching.
For \textit{StabStitch++}, it takes 35.3ms to deal with one frame, which is only a little longer than that of  \textit{StabStitch} \cite{nie2024eliminating}. Theoretically, \textit{StabStitch++} would double the inference time to generate bidirectional warps. However, due to the bidirectional decomposition module of the spatial warp model and the lightweight structure of the temporal warp model, the total running time only increases by 7.1ms.

\setlength{\tabcolsep}{4pt}
\begin{table}[t]
	\begin{center}
		\renewcommand{\arraystretch}{1.1}
		\caption{The effectiveness of the proposed bidirectional warps on alignment performance.}
        \vspace{-0.2cm}
		\label{ablation2}
		\begin{tabular}{cccc}
			\hline
			 & Method & UDIS-D \cite{nie2021unsupervised} & StabStitch-D \cite{nie2024eliminating}  \\
			\cline{2-4}
            1  & UDIS++ \cite{nie2023parallax}& 25.43/0.838  & 29.78/0.891  \\
            2  & UDIS++ (Bidirectional Warps)& $\mathbf{26.13}$/$\mathbf{0.852}$& $\mathbf{31.34}$/$\mathbf{0.907}$  \\
			\hline
		\end{tabular} 
	\end{center}
  \vspace{-0.5cm}
\end{table}

\begin{figure}[t]
	\centering
    
	\includegraphics[width=0.96\linewidth]{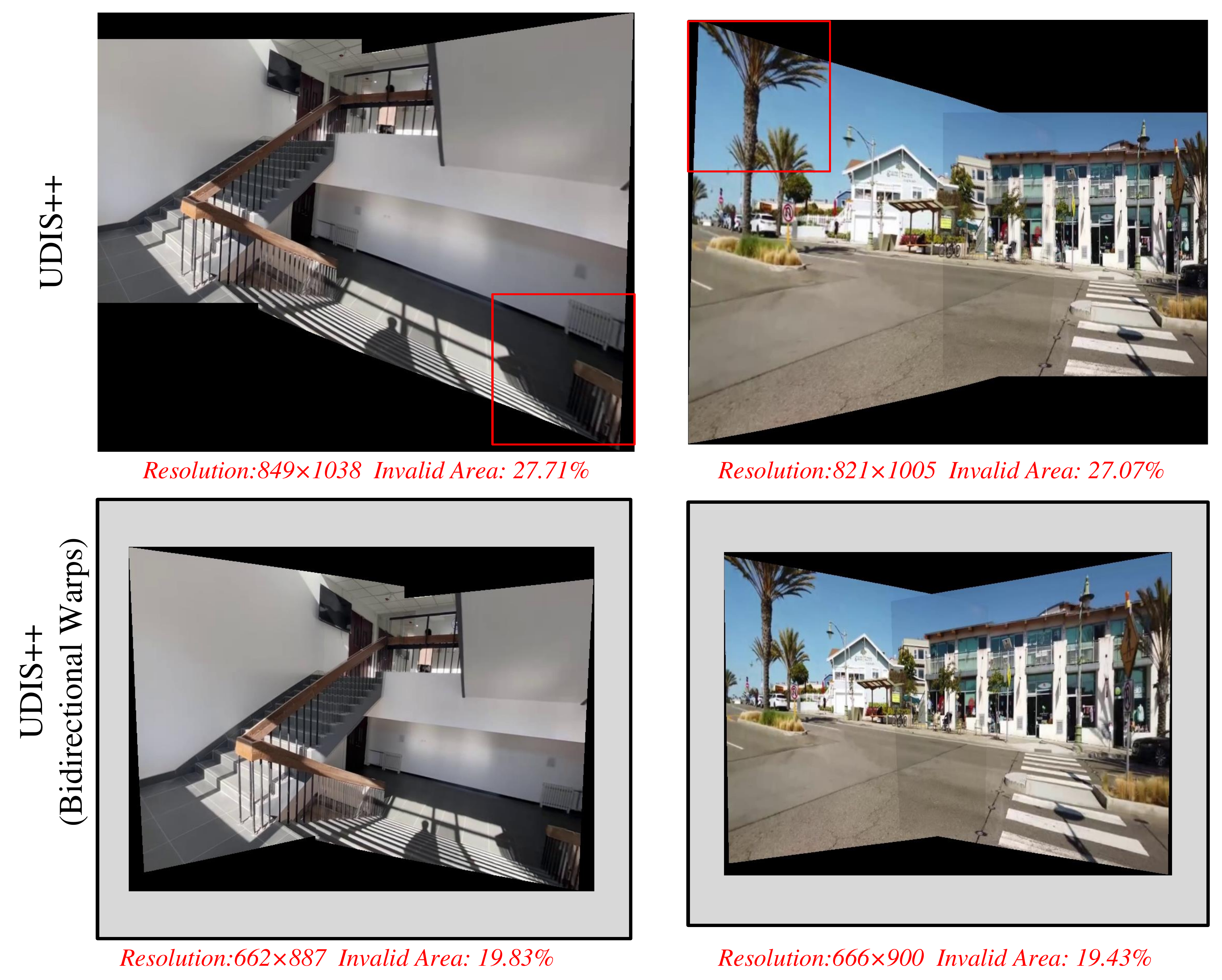}
 \vspace{-0.25cm}
	\caption{The effectiveness of the proposed bidirectional decomposition module on distortion performance. The red rectangles indicate projective distortions. The gray regions in the second row represent the space saved by projecting source views onto the middle plane.}
	\label{fig:biwarp}
\end{figure}

\subsection{Ablation Study}
Here, we conduct extensive ablation studies to reveal the effectiveness of the proposed solution.


\setlength{\tabcolsep}{4pt}
\begin{table*}[t]
	\begin{center}
		\renewcommand{\arraystretch}{1.1}
		\caption{Ablation studies on different objective terms.}
        \vspace{-0.2cm}
		\label{ablation results}
		\begin{tabular}{ccccccccccc}
			\hline
			 & $\mathcal{L}_{data}$ & $\mathcal{L}_{smooth}$ & $\mathcal{L}_{shape}$ & $\mathcal{L}_{online}$ & $\mathcal{L}_{trajectory}$ & $\mathcal{L}_{align}$ & Augmentation & Alignment $\uparrow$ & Stability $\downarrow$ & Distortion $\downarrow$  \\
			\cline{2-11}
            1 & \Checkmark & & &  &  & & & 30.30/0.893 & 51.78 & 0.338  \\
            2 & \Checkmark & \Checkmark &  & &  & & & 29.38/0.876 & 41.60  & 0.594  \\
            3 & \Checkmark & \Checkmark & \Checkmark &  & &  & & 28.83/0.869 & 41.57& 0.313 \\
            4 & \Checkmark &\Checkmark &\Checkmark & \Checkmark &  & & & 28.91/0.871 & $\mathbf{41.54}$ & $\mathbf{0.299}$  \\
            5 & \Checkmark &\Checkmark &\Checkmark & \Checkmark & \Checkmark & & & 28.24/0.857 & 41.81 & 0.319  \\
            6 & \Checkmark &\Checkmark &\Checkmark & \Checkmark & \Checkmark & \Checkmark & & 30.77/0.897 & 41.88 & 0.469  \\
            7 & \Checkmark &\Checkmark &\Checkmark & \Checkmark & \Checkmark & \Checkmark & \Checkmark & $\mathbf{30.88}$/$\mathbf{0.898}$ & 41.70 & 0.371  \\
			\hline
		\end{tabular} 
	\end{center}
  \vspace{-0.5cm}
\end{table*}

\begin{figure}[t]
	\centering
	\includegraphics[width=0.95\linewidth]{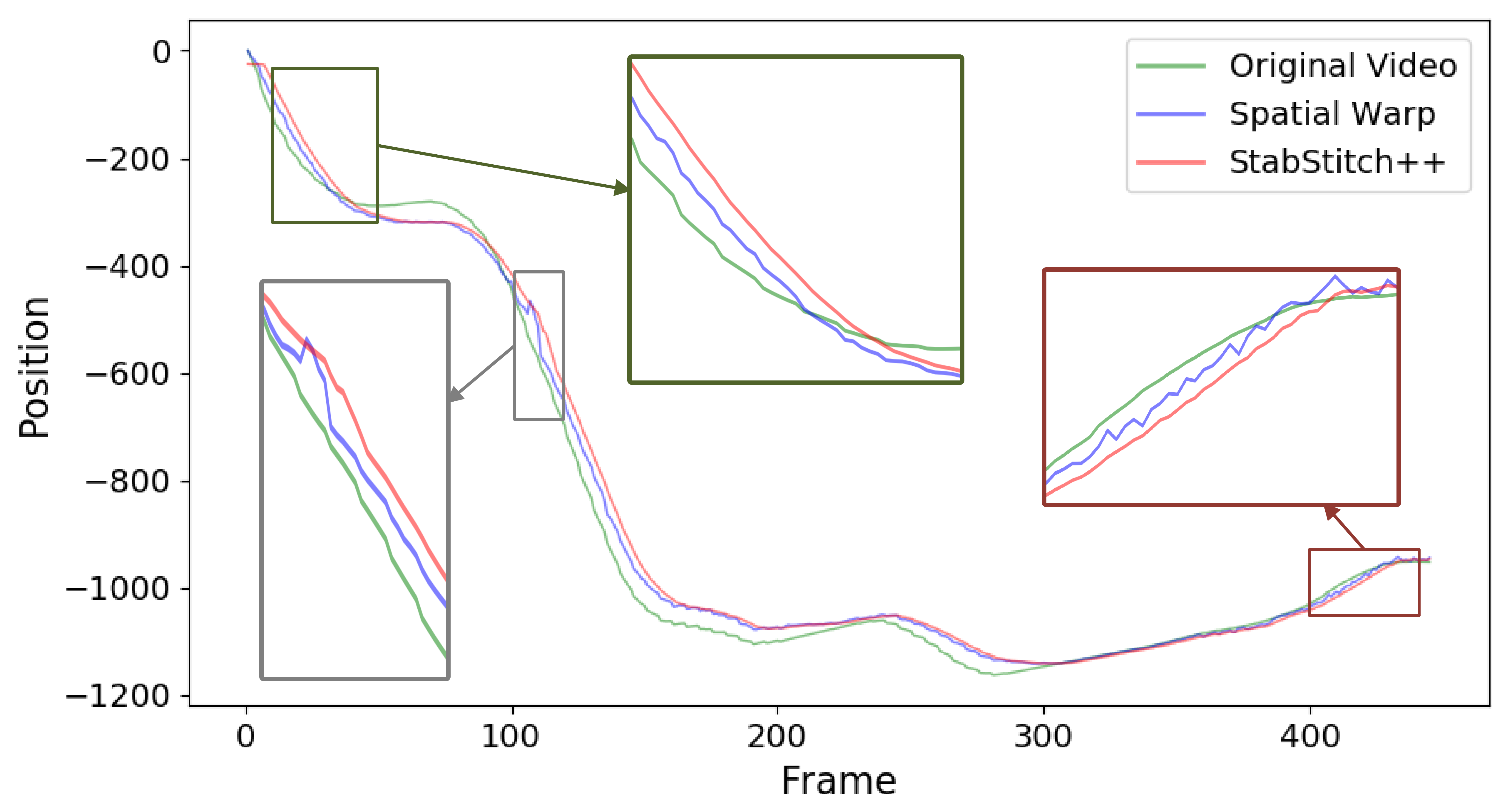}
 \vspace{-0.35cm}
	\caption{Trajectory visualization. It exhibits the complete trajectories of the original video and warped videos (by our spatial warp and \textit{StabStitch++}) to show the occurrence and elimination of warping shakes.}
	\label{fig:trajectory}
  \vspace{-0.3cm}
\end{figure}

\vspace{0.1cm}
\noindent\textbf{Bidirectional Warps:}
In addition to video stitching, the proposed bidirectional decomposition module can also be applied in the more common field (\textit{i.e.}, image stitching) and bring significant gains.
We demonstrate the benefit of alignment in Tab. \ref{ablation2}, where we validate its effectiveness on both image stitching (UDIS-D \cite{nie2021unsupervised}) and video stitching  (\textit{StabStitch-D} \cite{nie2024eliminating}) datasets. Moreover, we display the qualitative comparisons in Fig. \ref{fig:biwarp}, where we further report the output resolution and the rate of invalid areas. Compared with the unidirectional warp (\textit{i.e.}, UDIS++), the proposed bidirectional decomposition module evenly spreads projective distortions across both views, yielding more natural stitched results with smaller output resolutions and fewer invalid areas.

\vspace{0.1cm}
\noindent\textbf{Objective Terms:}
\textit{StabStitch++} is an unsupervised framework that is trained with a hybrid loss. Hence, we evaluate the effect of each objective term, and the results are demonstrated in Tab. \ref{ablation results}. (Experiment 1) When there is only the data term, \textit{StabStitch++} degenerates into an ordinary image stitching model. (Experiment 2) Then, we add the smoothness term. With their joint action, the stability of stitched videos improves but alignment and distortion worsen. (Experiment 3) The shape-preserving term significantly preserves the natural structures but further decreases the alignment performance. (Experiment 4) Next, we introduce the online collaboration term, which ensures the warping consistency among the sliding windows and brings comprehensive improvements. (Experiment 5) Compared with \textit{StabStitch} \cite{nie2024eliminating}, we design a trajectory consistency for the bidirectional warps from different views. Although the metrics are slightly decreased, it ensures the consistency of two views after warping. (Experiment 6) One of the biggest differences with \textit{StabStitch} \cite{nie2024eliminating} is that the proposed \textit{StabStitch++} can simultaneously optimize the alignment and stabilization. With the online alignment term, \textit{StabStitch++} can find the optimal solution that satisfies both conditions in the online mode rather than sacrificing one condition to enhance another. (Experiment 7) Finally, we validate the effect of our data augmentation strategy (mentioned in Sec. \ref{details}). Similar to the online collaboration term, it also brings in comprehensive improvements including alignment, stability, and distortion.


\vspace{0.1cm}
\noindent\textbf{Trajectory Visualization:}
We visualize the camera trajectories of the original video and the stitching trajectories of warped videos from the testing set of \textit{StabStitch-D} dataset in Fig. \ref{fig:trajectory}. Here, the trajectories are extracted from a certain control point in the horizontal direction. It can be observed that even if the input video is stable, image stitching can introduce undesired warping shakes, whereas \textit{StabStitch++} minimizes these shakes as much as possible during stitching.

%% file: sections/Limitation.tex
\section{Limitation}
\label{sec:Limitation}

While \textit{StabStitch++} demonstrates impressive performance across diverse scenarios, several limitations remain open for future exploration.

Although our method uses semantic features to improve robustness in low texture compared to traditional feature-based approaches, alignment quality may degrade when overlapping regions lack sufficient visual cues (e.g., uniform walls or skies). This is a common challenge for vision-based methods relying on appearance consistency. In the future, the pre-trained foundational models or other geometric priors may be leveraged to infer structural information in such cases, potentially improving alignment in feature-sparse environments.
Besides, to ensure computational efficiency, \textit{StabStitch++} processes downscaled versions of high-resolution videos to estimate spatiotemporal transformations, which are then upscaled to the original resolutions and applied to the source videos. While this approach avoids excessive GPU resource consumption, it may introduce precision loss during the scaling process, particularly for ultra-high-resolution inputs (e.g., 4K or beyond). Future work could explore dynamic resolution adjustments, where the system processes high-resolution videos at lower scales for efficiency and selectively refines critical regions at full resolution for accuracy.

%% file: sections/Conclusion.tex
\section{Conclusion}
\label{sec:Conclusion}

In this paper, we retarget video stitching to an emerging issue, named \textit{warping shake}, considering the current development of stabilization technology. It describes the undesired content instability caused by temporally unsmooth warps when image stitching technology is directly applied to videos. 
To solve this problem, we present \textit{StabStitch++}, an unsupervised online video stitching framework with spatiotemporal bidirectional warps. Concretely, we first estimate spatial bidirectional warps by determining the virtual middle plane and then integrate them with temporal warps to formulate the mathematical expression of stitching trajectories. Next, a warp smoothing model is proposed to simultaneously achieve content alignment, trajectory smoothness, and online collaboration using a hybrid objective loss.
Moreover, a video stitching dataset with various camera motions and scenes is built, which we hope can work as a benchmark and promote other related research work. 
Compared with our conference version \cite{nie2024eliminating}, we extend \textit{StabStitch} \cite{nie2024eliminating} to \textit{StabStitch++} through bidirectional decomposition and joint optimization, yielding better alignment, fewer distortions, and higher stability.
